\documentclass[sigconf]{acmart}
\usepackage{booktabs} 
\newcommand{\eat}[1]{}

\begin{document}
\title{Active Learning for Graph Embedding}

\begin{abstract}
Graph embedding provides an efficient solution for graph analysis by converting the graph into a low-dimensional space which preserves the structure information. In contrast to the graph structure data, the i.i.d. node embeddings can be processed efficiently in terms of both time and space. Current semi-supervised graph embedding algorithms assume the labelled nodes are given, which may not be always true in the real world. While manually label all training data is inapplicable, how to select the subset of training data to label so as to maximize the graph analysis task performance is of great importance. This motivates our proposed active graph embedding (AGE) framework, in which we design a general active learning query strategy for any semi-supervised graph embedding algorithm. AGE selects the most informative nodes as the training labelled nodes based on the graphical information (i.e., node centrality) as well as the learnt node embedding (i.e., node classification uncertainty and node embedding representativeness). Different query criteria are combined with the time-sensitive parameters which shift the focus from graph based query criteria to embedding based criteria as the learning progresses. Experiments have been conducted on three public datasets and the results verified the effectiveness of each component of our query strategy and the power of combining them using time-sensitive parameters. Our code is available online\footnote{\small{https://github.com/vwz/AGE}}.
\end{abstract}

%
%



\eat{\author
{Hongyun Cai{\small $~^{{\dagger}{\ddagger}}$}, Vincent W. Zheng{\small $~^{{\dagger}}$}, Kevin Chen-Chuan Chang{\small $~^{\diamond}$} }%
\vspace{1.6mm}\\
\affiliation{$^{\dagger}$\, Advanced Digital Sciences Center, Singapore}  \vspace{1mm} \\
\affiliation{$~^{\#}$\, Zhejiang University City College, China} \hspace{1.8cm}
\affiliation{$^{\diamond}$\, University of Illinois at Urbana-Champaign, USA}  \vspace{1mm} \\
\affiliation{\{hongyun.c, vincent.zheng\}@adsc.com.sg, zhufanwei@zju.edu.cn, kcchang@illinois.edu, huang@itee.uq.edu.au} \vspace{5mm} \\}

\eat{\author{Hongyun Cai}
\affiliation{%
  \institution{Advanced Digital Sciences Center, Singapore}}
\email{hongyun.c@adsc.com.sg}

\author{Vincent W. Zheng}
\affiliation{%
  \institution{Advanced Digital Sciences Center, Singapore}}
\email{vincent.zheng@adsc.com.sg}

\author{Kevin Chen-Chuan Chang}
\affiliation{%
  \institution{University of Illinois at Urbana-Champaign, USA}}
\email{kcchang@illinois.edu}}

\author{Hongyun Cai {\small $~^{{\dagger}}$}, Vincent W. Zheng {\small $~^{{\dagger}}$}, Kevin Chen-Chuan Chang {\small $~^{\diamond}$}}
\affiliation{%
  \institution{{\small $~^{{\dagger}}$} Advanced Digital Sciences Center, Singapore\, {\small $~^{\diamond}$}University of Illinois at Urbana-Champaign, USA}}
\email{hongyun.c@adsc.com.sg, vincent.zheng@adsc.com.sg, kcchang@illinois.edu}

\maketitle

\section{Introduction}
Nowadays graph (or network) is becoming more and more popular in many areas, e.g., citation graph in the research area, social graph in social media networks and so on. Directly analysing these graphs may be both time consuming and space inefficient. One fundamental and effective solution is graph embedding, which embeds a graph into a low-dimensional space that preserves the graph structure and other inherent information. With such kind of node representations, the graph analytic tasks, such as node classification, node clustering, link prediction, etc., can be conducted efficiently in both time and space \cite{hope}.

\eat{Graph embedding focuses on embedding a graph into low-dimensional space in which the graph structure is preserved. It has attracted researchers' interests in recent decades mainly for two reasons. First, converting graph to a set of node embeddings can benefit a lot of graph analysis application, such as node classification, node clustering (or community detection) and so on. More specifically, once the nodes are embedded into a space which well preserves the structure information, a large amount of state-of-the-art algorithms designed for non-structural data can now be applied on the node embeddings for graph analysis. Second, graph embedding algorithms have shown to outperform many other graph analysis algorithms which directly classify (or cluster) the nodes in graph. (!!! need some references here!!!)

[motivation of graph embedding: 1. graph data is very popular, how to learn graph data representation is critical as we can conduct the analytic tasks efficiently in both time ans space (from asymmetric transitive preserving graph embedding)]}

The graph embedding algorithms can be divided into two categories based on whether the label information is involved in the training: unsupervised and semi-supervised. In this work, we focus on the latter. Due to the success of deep learning in different areas, the latest semi-supervised graph embedding algorithms (e.g., \cite{planetoid, kipf2016semi}) devote to design a neural network model to embed the nodes. However, these methods assume the training labelled data is given which may not be always true in the real world. Take Twitter as an example, for a twitter network graph in which each node represents a user and a link represents the following relationships between two users, the node label can be different user attributes such as occupation, interest and so on. Manually labelling all users for training is inapplicable. To embed such a Twitter graph, we need a certain amount of users with label information for training. Obviously, different sets of training labelled nodes will lead to different graph embedding performance. Given a labelling budget, how to select the training labelled nodes so as to maximize the final performance is thus of great importance. Active learning (AL) is proposed to solve such kind of problems. 

Given a labelling budget, our objective is to design an active graph embedding framework which optimizes the performance of semi-supervised graph embedding algorithms by actively selecting the training labelled nodes. There are two main challenges for active graph embedding. First, different from traditional AL algorithms which are designed for independent and identically distributed data, the active graph embedding should consider the graph structure when select the ``informative'' nodes to label. Second, there are two major components in an active graph embedding framework: the active learning component and the graph embedding component. How to combine these two processes to make them reinforce each other so as to maximize the performance is non-trivial. 

In this paper, we proposed an effective \textbf{A}ctive \textbf{G}raph \textbf{E}mbedding (AGE) framework which tackles the above mentioned challenges. Specifically, we consider two popular AL query criteria, uncertainty and representativeness, to select the most informative node to label. For uncertainty, an information entropy score is calculated. For representativeness, in addition to the information density score which is widely adopted in most AL algorithms, we also propose a graph centrality score which calculates the PageRank centrality of each node to evaluate its representativeness. All the three informativeness scores are combined linearly. The query of the active learning process is raised at the end of every epoch of the graph embedding training process. As the process progresses, the graph embedding will generate more and more accurate node embedding as more informative labelled nodes are provided for training. Meanwhile, with the more accurate node embedding, the AL query strategy is able to find the more and more informative nodes because both the uncertainty and information density scores are calculated based on the embedding results. Moreover, considering these two scores are based on node embeddings which may be not correct at the beginning of the training, inspired by \cite{altextre} we combine the three AL scores with the time-sensitive parameters which give higher weight to graph centrality at the beginning and shift the focus to the other two scores as the training process progresses. In this paper, we use GCN as an example graph embedding algorithm. Our AGE framework can be directly applied on any other graph embedding algorithms. More details of AGE framework and our proposed AL criteria are introduced in Section \ref{sec:age}.

The contributions of this paper are summarized as below:
\begin{itemize}
\item To the best of our knowledge, we are the first to propose an active graph embedding framework which optimizes the graph embedding performance by actively selecting the labelled training nodes.
\item We defines three node informativeness criteria including uncertainty, information density and graph centrality, and extensively study the impact of them on active graph embedding performance.
\item We conduct comprehensive experiments on three public citation datasets. The results prove the superiority of our proposed AGE framework over other AL algorithms and a pipeline baseline.
\end{itemize}

The rest of this paper is organized as follows. We review the literature related to graph embedding and active learning in Section \ref{sec:rw}. Section \ref{sec:pl} introduces the example graph embedding algorithm GCN and the problem to solve in this paper. Our proposed active graph embedding algorithm is elaborated in Section \ref{sec:age}, followed by the experiment results analysis in Section \ref{sec:exp}. Finally, we conclude the paper in Section \ref{sec:cl}.

\section{Related Work}
\label{sec:rw}
In this paper, we focus on actively selecting labelled training instances so as to maximize the graph embedding performance with limited labelling budget. In this section, we review the literature in two relevant topics: graph embedding and active learning.
\subsection{Graph Embedding}
Graph embedding aims to embed the graph into low-dimensional space which preserves the graph structure information. The earlier studies \cite{TenenbaumEtAl2000, NIPS2001_1961, Roweis00nonlineardimensionality} tend to first construct the affinity graph based on feature similarity and then solve the eigenvector of the affinity graph as the node embeddings. These methods usually suffer from the high computational cost.  Recently, some graph embedding studies (e.g., LINE \cite{line}, GraRep \cite{grarep}) carefully designed the objective functions to preserve the first-order, second-order and/or high-order proximities. However, both LINE and GraRep are sub-optimal as the embeddings are separately learnt for different $k$-step neighbours. \eat{HOPE \cite{hope} considers to preserve asymmetric transitivity in directed graphs and proposes a scalable algorithm to approximate high-order proximities.}Motivated by the recent success of deep learning, some researchers start to learn the node embedding using the deep models. A part of them use truncated random walk (e.g., DeepWalk\cite{deepwalk}) or biased random walk (e.g., Node2vec \cite{node2vec}) to sample paths from graphs, and then apply skip-gram on the sampled paths so as to preserve the second-order proximities. In contrast to those who adopt skip-gram from language model, SDNE \cite{sdne} proposed a new deep model which jointly optimize first-order and second-order proximity and address the high non-linearity challenge. We aim to actively select the training labelled data so as to optimize the learnt embeddings given a fixed labelling budget. All the above methods are not applicable to our settings as they are unsupervised learning. The graph-based semi-supervised learning usually defines the loss function as a weighted sum of the supervised loss over labelled instances (denoted as $L_s$) and a graph Laplacian regularization term (denoted as $L_u$). $L_s$ is the standard supervised loss function such as squared loss, log loss or hinge loss. Various $L_u$ have been designed in the literature to incur a large penalty when connected nodes with large edge weight are predicted to have different labels \cite{Talukdar:2009, Belkin:2006:MRG:1248547.1248632}, or different embeddings \cite{Weston2012}. Planetoid \cite{planetoid} propose a feed-forward neural network framework and format $L_u$ as the log loss of predicting the context using the node embedding. However, the graph Laplacian regularization relies on the assumption that connected nodes in the graph are likely to share the same labels. This assumption may be not always true as graph edges could indicate other information in addition to node similarity. In observation of this, \cite{kipf2016semi} proposed GCN to encode the graph structure directly using a neural network model and train on a supervised loss function for all nodes, thus avoiding explicit graph-based regularization in the loss function. GCN has shown its superiority by outperforming the other state-of-the-art algorithms in terms of node classification. We adopt GCN as an example graph embedding framework in this work and will introduce more about it in Sect. \ref{sect:gcn}.

\subsection{Active Learning}
\begin{table*}[t]
\caption{Comparison between different active learning query strategies.}
\label{tab:comparedmethods}
\centering
\tabcolsep=0.05cm
\begin{small}
\begin{tabular} { | c|c |l|l|l|} \hline
\textbf{Categories}& \textbf{Sub-categories} &\textbf{Main idea} & \textbf{Strong points} & \textbf{Weaknesses} \\\hline
 & Uncertainty Sampling \cite{alsfslt}& Label the most uncertain instances & & \\ \cline{2-3}
  &Query-by-Committee \cite{bilgic:icml10}&Label the instances that multiple classifiers &Simple and fast approaches  &May find the noisy and \\ 
  Heterogeneity based&&disagree most&to identify the most &unrepresentative regions\\\cline{2-3}
  &Expected Model & Label the instances which are most different &  unknown regions&  \\ 
  &Change \cite{altextre}&from the current known model&& \\\hline
  & Expected Error & Minimize label uncertainty of the remaining &Directly optimize the & Too expensive to compute  \\ 
  Performance based &Reduction \cite{oalmi}& unlabelled instances&model performance&all unlabelled data \\\cline {2-5}
  
  & Expected Variance & The variance typically reduces when the & Efficiently express model    &  Only applicable to limited models, e.g.,\\ 
  &Reduction \cite{Schein2007}& error of the model reduces& variances in closed form &  neural networks, mixture models \\\hline
  Representativeness  &   & Label the instance that can represent the & Avoid outlier by the & Not informative enough, usually\\ 
  based \cite{mlsvm}& &  underlying distribution of training instances&representativeness component & combined with other criteria\\\hline
  
\end{tabular}
\end{small}
\end{table*}
In many domains, labelled data is often expensive to obtain. Active learning (AL) is thus proposed to train a classifier that accurately predicts the labels of new instances while requesting as few training labels as possible \cite{alsurvey}. An AL framework usually consists of two primary components: a \textbf{query system} which picks an instance from the training data to query its label and an \textbf{oracle} which labels the queried instance. Researchers have proposed various algorithms to optimize the training performance given a fixed labelling budget. Based on the query strategy, the majority work can be divided into three categories \cite{alsurvey}: the heterogeneity based, the performance based and the representativeness based. The detailed comparisons between them are listed in Table \ref{tab:comparedmethods}. Generally speaking, different implementations of the three major AL categories can be proposed for different classification algorithms. There does not exist an ``optimal'' AL solution for all classification tasks. Our active graph embedding is distinct from the most AL algorithms in two ways: the training instances are in graph structure rather than \textit{i.i.d.}, and representation of training nodes are learnt during the classifier training process instead of being given as the fixed input. On one hand, several attempts have been made for AL in graph \cite{bilgic:icml10, ssgc}, in which the graph structure are utilized to train the classifiers and/or calculate the AL query scores when selecting the node to label. Compared with them, we utilize not only the graph structure but also the embeddings learnt during training process to select the informative nodes to label. Moreover, they do not learn any node embeddings but just simply graph classification. On the other hand, only limited work (i.e., \cite{altextre}) has been done to consider AL strategies for instance representation learning algorithm. In \cite{altextre}, the authors proposed to select the examples that are likely to affect the representation-level parameters (embeddings) for text classification with embeddings. Their algorithm is specifically designed to the classification model which has embeddings as the model parameters and thus is not applicable to more general graph embedding work such as GCN in our framework.

\section{Preliminary}
\label{sec:pl}
The notations used in this paper are summarized in Table \ref{tab:notations}. Next we introduce more about the semi-supervised graph embedding algorithm we adopted in our framework, i.e., GCN \cite{kipf2016semi}.
\begin{table}[t]
\caption{Notations}
\label{tab:notations}
\centering
\tabcolsep=0.05cm
\begin{small}
\begin{tabular} { | l|l |} \hline
\textbf{Notations}& \textbf{Descriptions} \\ \hline
$\mathcal{G}$ = $(\mathcal{V},\mathcal{E})$& Graph $\mathcal{G}$ with nodes set $\mathcal{V}$ and edges set $\mathcal{E}$\\ \hline
$A$, $D$ & The adjacent matrix and degree matrix of $\mathcal{G}$ \\ \hline
$X$ & Node feature matrix, each row corresponds to the feature \\
& vector of a node in $\mathcal{G}$ \\ \hline
$N$, $C$ & Number of nodes, classes in $\mathcal{G}$ \\ \hline
$F$ & Feature dimensionality of a node in $\mathcal{G}$  \\ \hline
$\mathcal{L}$, $\mathcal{U}$ & The set of labelled and unlabelled nodes \\ \hline
$L$, $U$ & The number of labelled and unlabelled nodes \\ \hline
$Y_{ic}$ & The indicator of node $v_i$ containing label $c$ \\ \hline
$Z_{ic}$ & The probability of node $v_i$ containing label $c$ predicted \\
& by GCN \\ \hline
$H^{(l)}$, $W^{(l)}$ & The matrix of activations and the trainable weight matrix in \\
& the $l$-th layer of GCN. \\ \hline

\end{tabular}
\end{small}
\end{table}
\subsection{GCN}
\label{sect:gcn}
Given a graph $\mathcal{G}$ = $(\mathcal{V},\mathcal{E})$ with $N$ nodes $v_i \in \mathcal{V}$, edges $(v_i,v_j) \in \mathcal{E}$, an adjacency matrix $A \in \mathbb{R}^{N \times N}$ (binary or weighted), a degree matrix ${D_{ii}} = \sum\nolimits_j {{A_{ij}}} $, a node feature matrix $X \in \mathbb{R}^{N \times F}$ (i.e., $F-$dimensional feature vector for $N$ nodes), label matrix for labelled nodes $Y \in \mathbb{R}^{L \times C}$ (i.e., $Y_{ij} = 1$ indicates node $i$ has label $j$), \cite{kipf2016semi} proposes a multi-layer Graph Convolutional Network (GCN) for semi-supervised node classification on $\mathcal{G}$. Unlike traditional graph-based semi-supervised learning which assumes that the connected nodes are likely to share the same labels, GCN avoids such kind of explicit graph-based regularization in the loss function by encoding the graph structure directly using their proposed neural network model.

Specifically, the layer-wise propagation rule of GCN is defined as:
\begin{equation}
\label{eq:lwpro}
H^{(l+1)} = \sigma(\tilde{D}^{-\frac{1}{2}}\tilde{A}\tilde{D}^{-\frac{1}{2}}H^{(l)}W^{(l)}).
\end{equation}
where $\tilde{A}=A+I_N$ is the adjacency matrix of $\mathcal{G}$ with added self-connections. $I_N$ is the identity matrix and ${\tilde{D}_{ii}} = \sum\nolimits_j {{\tilde{A}_{ij}}} $. The active function $\sigma(\cdot)$ is defined as $ReLU(\cdot) = max(0,\cdot)$ for all layers expect for the output layer in their work. $W^{(l)}$ and $H^{(l)}$ denotes trainable weight matrix and the matrix of activations in the $l$-th layer respectively. $H^{(0)} = X$.

To train a GCN model ($f(X,A)$) with $M$ layers, $\hat{A}=\tilde{D}^{-\frac{1}{2}}\tilde{A}\tilde{D}^{-\frac{1}{2}}$ is first calculated in the pre-processing step. The $M$-th layer (the output layer) takes the following form:
\begin{equation}
\label{eq:z}
Z=f(X,A)=softmax(\hat{A}H^{(M-1)}W^{(M-1)})
\end{equation}
where $H^{(M-1)}$ is derived from Eq. \ref{eq:lwpro}, and $W^{(M-1)}$ is a hidden-to-output weight matrix. The active function in the last layer is $softmax(x_i)=\frac{1}{\Lambda} exp(x_i)$ with $\Lambda =\sum\nolimits_i {exp(x_i)}$, and it is applied row-wise.

Finally the supervised loss function is defined as the cross-entropy error over all labelled nodes:
\begin{equation}
\label{eq:loss}
\mathsf{Loss}=-\sum\limits_{l\in\mathcal{Y}_L}{\sum\limits_{c=1}^{C}Y_{lc}\ln {Z_{lc}}}
\end{equation}
where $\mathcal{Y}_L$ is the set of indices for labelled nodes. $Z$ is derived from Eq. \ref{eq:z}.

\subsection{Problem Formulation}
The input of the active graph embedding problem includes a graph $\mathcal{G}$ = $(\mathcal{V},\mathcal{E})$, along with its adjacency matrix $A$, its degree matrix $D$, its node feature matrix $X$, an oracle to label a query node, and a labelling budget $B$. Among the $N$ nodes $v \in \mathcal{V}$, $L$ nodes are initially labelled. Denote the set of labelled nodes as $\mathcal{L}$, and the unlabelled nodes set as $\mathcal{U}$.                                                                                                                                                                                                                                                                                                                                                                                                                                                                                                                                                                                                                                                                                                                                                                                                                                                                                                                                                                                                                                                                                                                                                                                                                                                                                                                                                                                                                                                                                                                                                                                                                                                                                                                                                                                                                                                                                                                                                                                                                                                                                                                                                                                                                                                                                                                                                                                                                                                                                                                                                                                                                                                                                                                                                                                                                                                                                                                                                                                                                                                                                                                                                                                                                                The objective of this work is to optimize the performance of the semi-supervised graph embedding algorithm (we use GCN introduced in Sect. \ref{sect:gcn} as an example in this work) by designing an active learning query strategy to select $B$ nodes from $\mathcal{U}$ for the oracle to label and add to $\mathcal{L}$ for graph embedding training.

\section{Active Graph Embedding}
\label{sec:age}
Given a fixed labelling budget, we propose an Active Graph Embedding (AGE) method to actively select the labelled training instances for optimizing graph embedding performance. Next, we introduce the details of our proposed AL strategy for graph embedding methods. Note that our proposed AL strategy can be applied to any semi-supervised graph embedding algorithm. In this work, we adopt the state-of-the-art algorithm GCN as an example graph embedding method for illustration. 
\subsection{Active Graph Embedding Framework}
\begin{figure}[t]
  \centering
    \includegraphics[width=\linewidth]{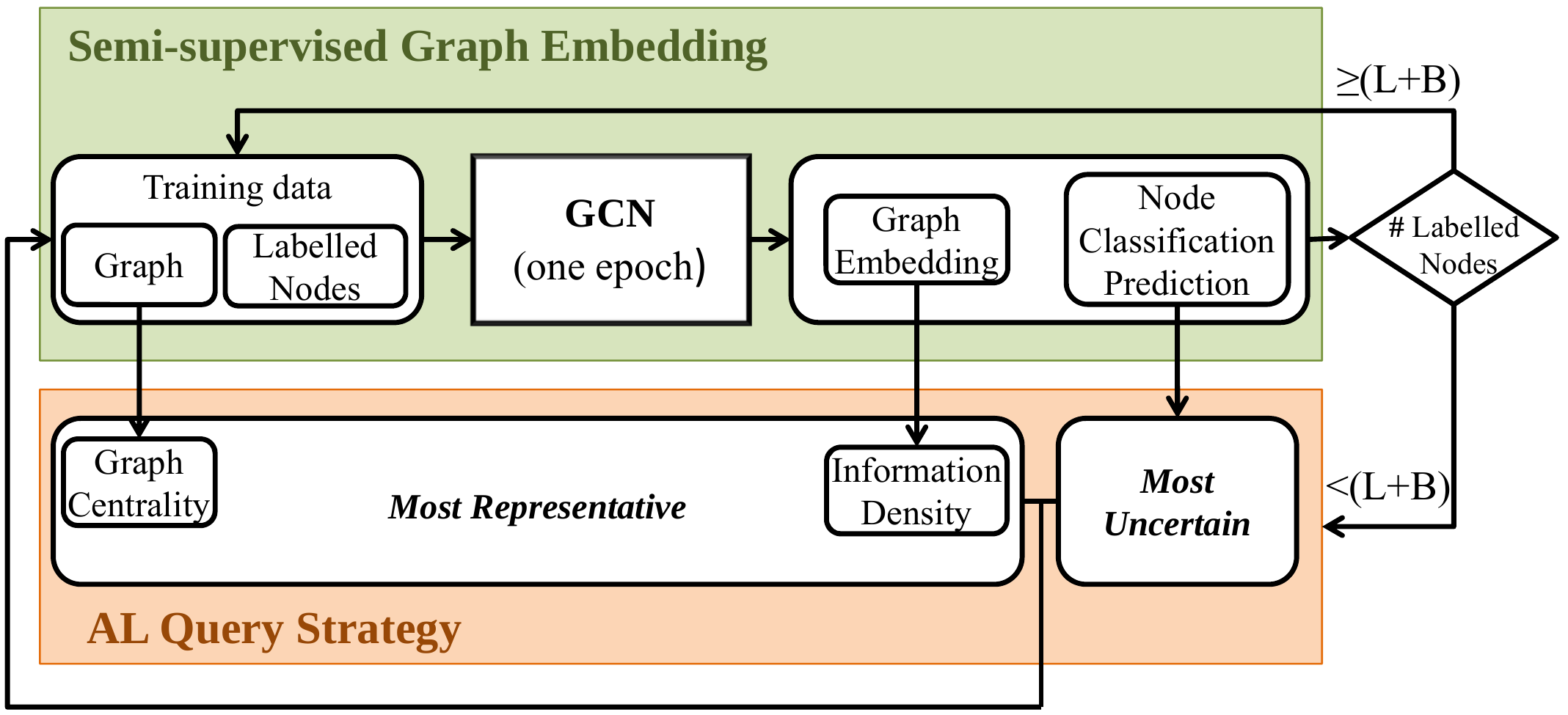}
  \caption{Framework of AGE.}
  \label{fig:framework}
\end{figure}
The framework of our proposed Active Graph Embedding (AGE) method is illustrated in Fig. \ref{fig:framework}. AGE takes a graph $\mathcal{G}$ = $(\mathcal{V},\mathcal{E})$ and a small set of initial labelled nodes as input. GCN \cite{kipf2016semi} is then applied on the training data for graph embedding and node classification. At the end of every epoch of GCN, AGE will check whether the labelling budget $B$ is reached. If yes, another training epoch of GCN will be processed directly. Otherwise, our proposed AL query strategy will pick one(or a few in the batch mode) best candidate(s) from all unlabelled nodes ($\mathcal{U}$), ask the oracle to label it, and put it in the labelled nodes set ($\mathcal{L}$). Then another epoch of GCN will be trained on the updated training data with the newly added labelled node. This procedure will be repeated until GCN converges.

Here a question arises: what is the best candidate node to label at each iteration? We follow most AL studies and select two widely adopted AL query criteria, i.e., uncertainty (e.g., \cite{alsfslt}) and representativeness (e.g., \cite{mlsvm}), in our proposed AL query strategy. Next, we introduce how we define the uncertainty and representativeness in this work, as well as how we combine these two criteria in one objective function.

\subsection{Active Learning Criteria}
\subsubsection{Uncertainty} As one of the most commonly used AL query strategy, uncertainty sampling queries the labels for the nodes which current model is least certain with represent to classification prediction. In this paper, we use the general uncertain measure, i.e., \textbf{information entropy} $\mathbf{\phi _{entropy}}$, as our informativeness metric. The information entropy of a candidate node $v_i$ is calculated as:
\begin{equation}
\label{eq:entropy}
{\phi _{entropy}}({v_i}) =  - \sum\limits_{c = 1}^C {P({Y_{ic}} = 1|\mathcal{G},\mathcal{L},X)\log P({Y_{ic}} = 1|\mathcal{G},\mathcal{L},X)} 
\end{equation}
where $P({Y_{ic}} = 1|\mathcal{G},\mathcal{L},X)$ is the probability of node $v_i$ belonging to class $c$ predicted by GCN, i.e. $Z_{ic}$ in Eq. \ref{eq:loss}. The larger $\phi _{entropy}({v_i})$ is, the more uncertain current model is regarding to $v_i$.

\subsubsection{Representativeness} \label{sec:rep} One drawback of uncertainty sampling based AL query strategy is that it may find the noisy and unrepresentative region as it tries to explore the most unknown regions of the data \cite{alsurvey}. Consequently, a representativeness-based AL criterion is often considered to be combined with the uncertainty sampling based method so as to find the most informative node to label. We consider two representativeness measurements: the \textbf{information density} $\mathbf{\phi _{density}}$ and the \textbf{graph centrality} $\mathbf{\phi _{centrality}}$. The first measurement aims to find a node which is ``representative'' of the underlying data distribution in the embedded space, while the second metric measures the nodes by their centralities in the graph. Next, we introduce these two methods one by one.

To find the nodes that locate dense regions of the training nodes in the embedded space, we calculate the \textbf{density} score for each candidate node $v_i$ by first applying Kmeans on the embeddings of all unlabelled nodes then compute the Euclidean distance between each node and its cluster center (i.e., the average distance to the nodes in the same cluster). The density score of node $v_i$ is calculated by converting the distance value to similarity scores:
\begin{equation}
\label{eq:density}
\phi _{density}(v_i)=\frac{1}{1+ED(Emb_{v_i},CC_{v_i})} 
\end{equation}
where $ED()$ is the Euclidean distance measurement, $Emb_{v_i}$ is the embedding of node $v_i$ and $CC_{v_i}$ is the center of the cluster that $v_i$ belongs to. The larger $\phi _{density}(v_i)$ is, the more representative $v_i$ is in the embedding space.

One characteristic that makes our AGE different from most other AL algorithms is that the input instances are not i.i.d., but connected with links. The graphical structure is then utilized to calculate another node representativeness score based on graph centrality. The graph centrality was first proposed in \cite{Newman:2010:NI:1809753} to reflect the node's sociological origin in social network analysis. Various metrics have been proposed to measure the centrality of a node, from the classic methods (e.g., degree centrality, closeness centrality\cite{gcentrality}) to the more recent eigenvector-based metrics (e.g., PageRank centrality \cite{pagerankcentrality}). In this work, we adopt \textbf{PageRank centrality} to calculate $\mathbf{\phi _{centrality}}$ because it outperforms others as shown in Sect. \ref{sec:graphcentrality}. The PageRank centrality of a candidate node $v_i$ is calculated as:
\begin{equation}
\label{eq:centrality}
\phi _{centrality}(v_i)= \rho \sum\limits_j {{A_{ij}}\frac{{{\phi _{centrality}}({v_j})}}{{\sum_k A_{jk}}}}  + \frac{{1 - \rho }}{N}
\end{equation}
where $\rho$ is the damping parameter.

\subsection{Combination of Different Criteria}
\subsubsection{Score Normalization}
The scores derived from different criteria are on an incomparable scale, thus we convert them into percentiles as in \cite{altextre}. Denote $\mathcal{P}_{\phi}(v,\mathcal{U})$ as the percentile of nodes in $\mathcal{U}$ which have smaller scores than node $v$ in terms of metric $\phi$. Then the objective function of our proposed AGE to select the node for labelling is defined as:
\begin{equation}
\label{eq:obj}
{\alpha} \cdot {\mathcal{P}_{entropy}}(v,\mathcal{U}) + {\beta} \cdot {\mathcal{P}_{density}}(v,\mathcal{U}) + \gamma \cdot {\mathcal{P}_{centrality}}(v,\mathcal{U})
\end{equation}
where $\alpha + \beta + \gamma = 1$. Our objective is to select a node $v \in \mathcal{U}$, which maximize the above objective function (Eq. \ref{eq:obj}).

\subsubsection{Time-sensitive Parameters}
Instead of predefining the parameters $\alpha$, $\beta$ and $\gamma$, we follow the settings in \cite{altextre} to draw the parameters as the time-sensitive random variables. More specifically, since $\phi_{entropy}$ and $\phi_{density}$ are calculated based on the GCN outputs (i.e., the node classification predictions and the node embeddings ), the parameters of these two metric (i.e., $\alpha$ and $\beta$) should be smaller at the beginning of the AL iterations because the outputs may be not very accurate in the first few epochs. In contrast, $\gamma$ (i.e., the parameter for $\phi_{centrality}$ which purely relies on graph structure) should be larger. As learning progresses, GCN runs more epochs with more labelled training data, the model can now pay more attention to $\phi_{entropy}$ and $\phi_{density}$. This time-sensitive parameter is realized by drawing the parameters from a beta distribution, e.g. $\gamma_t \sim Beta(1, n_t)$. $n_t$ increases as the number of AL iterations increases, which will draw $\gamma_t$ with larger expectation. In contrast, $\alpha_t$, $\beta_t$ can be drawn from a beta distribution $Beta(1, n'_t)$ in which $n'_t$ decreases as AL iterations increases. Finally, the $\alpha_t$, $\beta_t$ and $\gamma_t$ drawn at timestamp $t$ are normalized to sum up to 1.

\section{Experiments}
\label{sec:exp}
We design experiments to: 1) verify the model design of our proposed AGE framework; 2) compare AGE with the other AL baselines. For the first objective, we verify the design of AGE from two perspectives: the adoption of PageRank centrality as graph centrality metric and the time-sensitive parameters to combine different AL criteria. For the second objective, we compare AGE with different active learning baselines. 

We first introduce our the experimental setup, followed by the experimental results analysis.

\subsection{Set Up}
We follow the experimental setup in the state-of-the-art semi-supervised graph embedding methods \cite{kipf2016semi, planetoid}. 

\subsubsection{Datasets} All experiments are conducted on three public citation network datasets -- Citeseer, Cora and Pubmed. The three datasets contain a list of documents, each of which is represented by sparse bag-of-words feature vectors. The documents are connected by citation links, which are treated as undirected and unweighted edges. The statistics of these datasets are summarized in Table \ref{tab:datastat}. 

\begin{table}[t]
\caption{Dataset statistics}
\label{tab:datastat}
\centering
\tabcolsep=0.05cm
\begin{small}
\begin{tabular} { | l|r |r|r|r|r|} \hline
\textbf{Dataset}& \textbf{$\mathbf{\#}$ Nodes} &\textbf{$\mathbf{\#}$ Edges} & \textbf{$\mathbf{\#}$ Classes} & \textbf{Feature Dim.} & \textbf{Label Rate}\\ \hline
Citeseer & 3,327 & 4,732 & 6 & 3,703 & 0.036 \\ \hline
Cora & 2,708 & 5,429 & 7 & 1,433 & 0.052 \\ \hline
Pubmed & 19,717 & 44,338 & 3 & 500 & 0.003 \\ \hline

\end{tabular}
\end{small}
\end{table}

\subsubsection{Dataset division for training, validation and testing} For each dataset, we use 500 nodes for validation, 1000 nodes for testing and the remaining nodes for training. The test instances set are the same as in \cite{kipf2016semi, planetoid}. We randomly sample 500 nodes from the non-test nodes and fixed them as the validation set across all experiments to ensure that the performance variation in the experiments is due to different active learning query strategies. We repeat this process for ten times and test all experiments on all the ten validation sets separately. We follow the label rate (i.e., the number of labelled nodes that are used for GCN training divided by the total number of nodes) used in the existing work \cite{kipf2016semi, planetoid}, denoted as $L_{train} = 20 \times C$, where $C$ is the number of classes in each dataset. Then the label budget for active learning methods is $B = L_{train}-L$, where $L$ is the number of initially labelled nodes.

\subsubsection{Initial labelled nodes sampling} \label{sec:ilns}
The active learning framework takes a few labelled nodes at the very beginning stage. Follow the settings in the existing work (e.g., \cite{kipf2016semi, planetoid}) which consider the label balance across classes, we randomly sample $\frac{L}{C}$ nodes for each class from the non-test and non-validation nodes as the initially labelled nodes. We repeat the process for 200 times and report the average results for all experiments. In both both \cite{kipf2016semi} and \cite{planetoid}, the number of initially labelled nodes per class is set as 20, denoted as $L_{semi}$. In our experiment, we initially label 4 nodes for each class. For fair comparison, the budget of AL strategy is set as $B = L_{semi} - L$, so that the same amount of labelled nodes are used to train GCN for all methods. Note that each algorithm is tested 2000 times (10 (validation sets) $\times$ 200 (initial labelled sets)).

\subsubsection{Evaluation metrics} One common task to evaluate the graph embedding performance is node classification. In this work, we adopt MacroF1 and MicroF1 \cite{deepwalk}, two classic classification evaluation measurements, to asset the node classification performance.  

All experiments are conducted on Linux computers equipped with Intel(R) 3.50GHz CPUs and 16GB RAMs. 

\subsection{Comparison of Graph Centrality Metrics}
\label{sec:graphcentrality}
\begin{figure}[t]
  \centering
    \includegraphics[width=\linewidth]{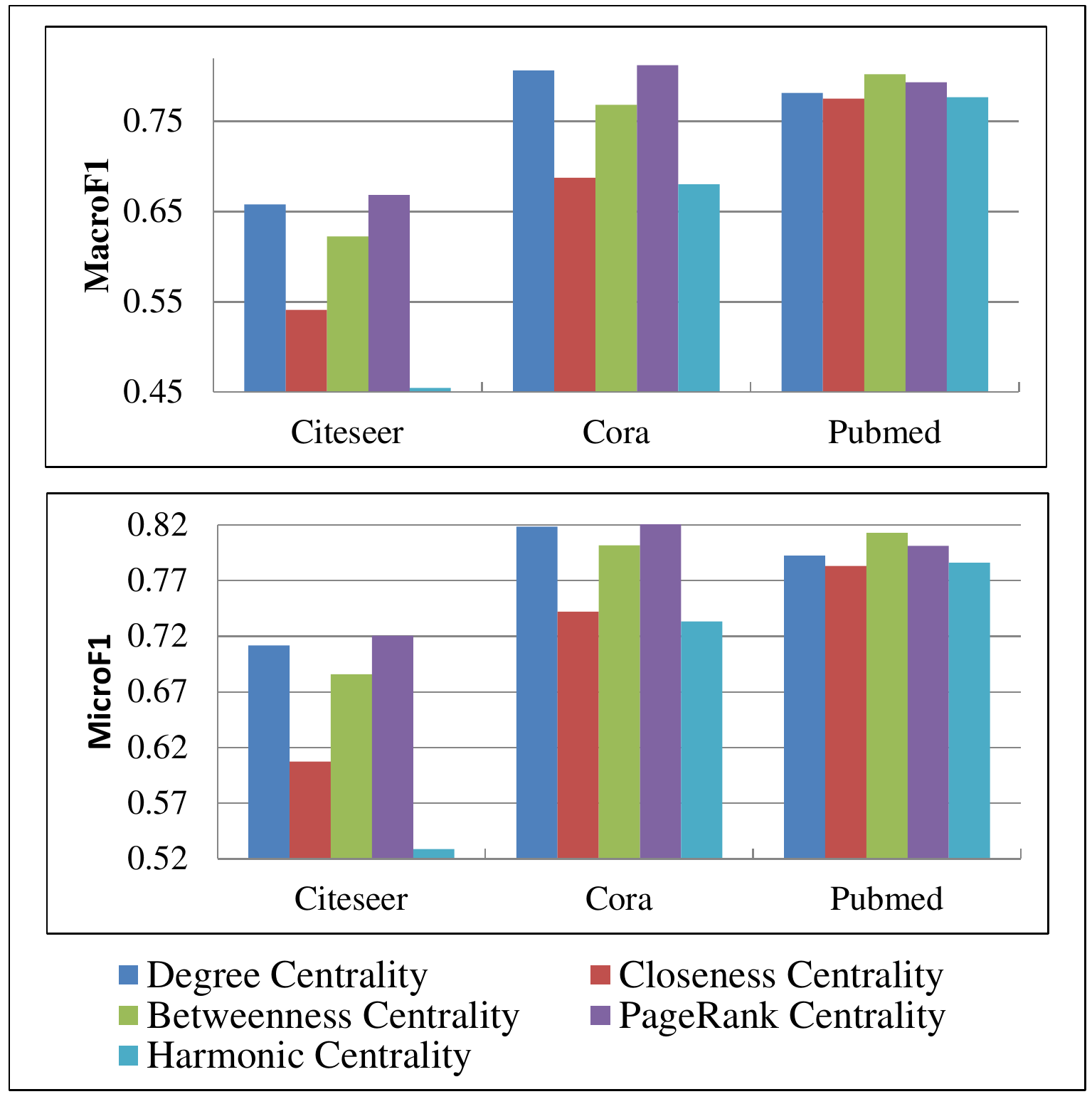}
  \caption{Comparison of Graph Centrality Metrics.}
  \label{fig:comparegc}
\end{figure}
\begin{figure*}[t]
  \centering
    \includegraphics[width=\linewidth]{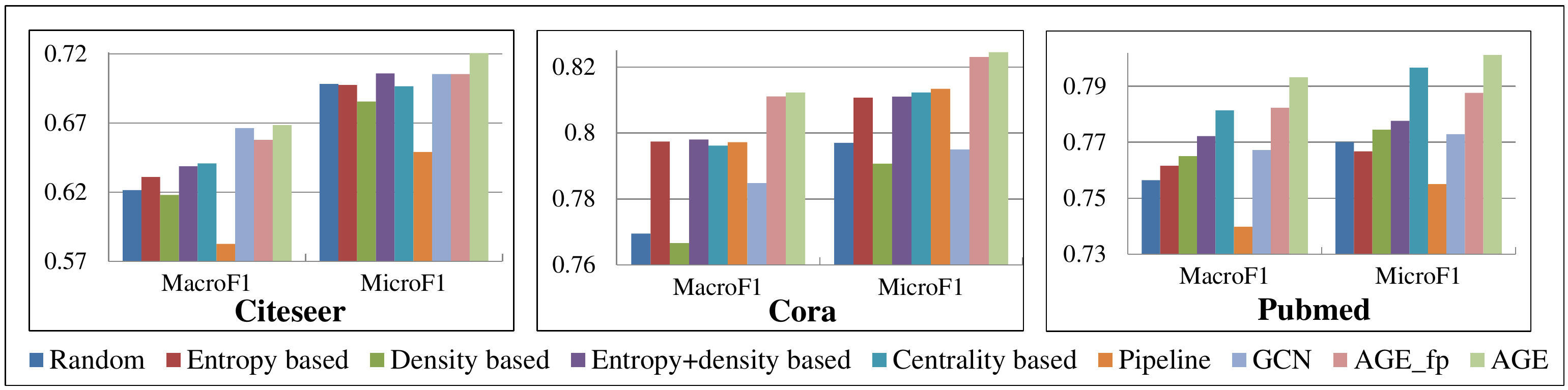}
  \caption{Comparison of AL Query Strategies in Terms of Node Classification Performance}
  \label{fig:compareal}
\end{figure*}
In this section, we first show the reason of adopting PageRank centrality (Eq. \ref{eq:centrality}) in our AGE framework. Five different graph centrality metrics are compared, including degree centrality, closeness centrality, betweenness centrality, PageRank centrality and harmonic centrality. All the five metrics are implemented using NetworkX \footnote{\small{https://networkx.github.io/}}. As shown in Fig. \ref{fig:comparegc}, PageRank centrality consistently outperforms the other centrality metrics on both Citeseer and Cora. Although betweenness centrality achieves slightly better performance on Pubmed, considering that PageRank outperforms betweenness on the other two datasets by 7.4\% (MacroF1) and 5.1\% (MicroF1) on Citeseer and 5.7\% (MacroF1) and 2.8\% (MicroF1) on Cora, we adopt PageRank as the graph centrality metric in our work.

\subsection{Effect of Time-sensitive Parameters}
\label{sec:ts}
Now we verify the superiority of using time-sensitive parameters over the predefined fixed parameters. For fair comparison, we tune the value of $\gamma$ with the other two parameters calculated as $\alpha = \beta = 0.5 \times (1-\gamma)$. The value of $\gamma$ is tuned within the range of $(0,1)$ with the step $0.1$. 
\begin{table}[t]
\caption{Effect of Time-sensitive Parameters}
\label{tab:ts}
\centering
\tabcolsep=0.05cm
\begin{small}
\begin{tabular} { | c|c|c|c|c|c|c|} \hline
{$\mathbf{\gamma}$}& \multicolumn{2}{|c|}{\textbf{Citeseer}} &\multicolumn{2}{|c|}{\textbf{Cora}} &\multicolumn{2}{|c|}{\textbf{Pubmed}} \\ \cline {2-7}
& \textbf{MacroF1} &\textbf{MicroF1} &\textbf{MacroF1} &\textbf{MicroF1}&\textbf{MacroF1} &\textbf{MicroF1} \\ \hline 
0.1 & 0.6576 & 0.7099 & 0.8040 & 0.8170 & 0.7717 & 0.7792 \\ \hline
0.2 & 0.6559 & 0.7050 & 0.8001 & 0.8142 &0.7721 & 0.7790 \\ \hline
0.3 & \underline{0.6601} & \underline{0.7067} & 0.802 & 0.8163 & 0.7751 & 0.7826 \\ \hline
0.4 & 0.6583 & 0.7037 & 0.8061 & 0.819 & 0.7704 & 0.7779 \\ \hline
0.5 & 0.6520 & 0.6975 & 0.8078 & 0.8201 & 0.7754 & 0.7819 \\ \hline
0.6 & 0.6503 & 0.6955 & 0.8096 & 0.8216 & 0.7747 & 0.7817 \\ \hline
0.7 & 0.6396 & 0.6847 & \underline{0.8111} & \underline{0.8230} & 0.7798 & 0.7864 \\ \hline
0.8 & 0.6399 & 0.6848 & 0.808 & 0.8210 & 0.7806 & 0.7870 \\ \hline
0.9 & 0.6416 & 0.6918 & 0.7993 & 0.8126 & \underline{0.7824} & \underline{0.7877} \\ \hline
time-sensitive & \textbf{0.6685} & \textbf{0.7206} & \textbf{0.8123} & \textbf{0.8245} & \textbf{0.7933}&\textbf{0.8012} \\ \hline

\end{tabular}
\end{small}
\end{table}

The results are illustrated in Table \ref{tab:ts} and the best results are highlighted in bold. Compared with deterministically setting the parameter for combining different active learning criteria, the time-sensitive parameters provide a more flexible way to find the balance between various criteria at different time, and thus show a better performance. We also underline the best results with predefined parameter for each dataset, i.e., Citeseer ($\gamma = 0.3$), Cora ($\gamma = 0.7$) and Pubmed ($\gamma = 0.9$). As shown in Table \ref{tab:ts}, compared with the fixed parameters setting, the time-sensitive parameters relatively improve the node classification performance 1.3\% (MacroF1) and 2\% (MicroF1) on Citeseer and 1.4\% (MacroF1) and 1.7\% (MicroF1) on Pubmed.
\subsection{Comparison of AL Query Strategies}

In this section, we compare with the following AL query strategies. For all strategies except for ``Random'' and ``Pipeline'', we randomly label $L$ nodes as introduced in Sect. \ref{sec:ilns}. Then during the training of GCN, we actively select $B$ nodes to label based on different AL metrics as shown in Fig. \ref{fig:framework}.
\begin{itemize}
\item Random: Randomly label $(L+B)$ nodes to train GCN.
\item Entropy based: Actively select nodes to label by Eq. \ref{eq:entropy}.
\item Density based: Actively select nodes to label by Eq. \ref{eq:density}.
\item Entropy+density based: Actively select nodes to label based on Eq. \ref{eq:entropy} $+$ Eq. \ref{eq:density}.
\item Centrality based: Actively select nodes to label by Eq. \ref{eq:centrality}.
\item Pipeline: Randomly label $L$ nodes to train GCN. After GCN converges, actively select $B$ nodes to label by Eq. \ref{eq:obj} with time-sensitive parameter $\alpha_t$, $\beta_t$ and $\gamma_t$. Finally, train GCN again with all $L+B$ labelled nodes. The pipeline approach is designed to verify that the AL process and graph embedding process can reinforce each other during the training.
\item AGE$\_$fp: Actively select nodes to label by Eq. \ref{eq:obj} with fixed (tuned) parameter $\alpha$, $\beta$ and $\gamma$.
\item AGE: Actively select nodes to label by Eq. \ref{eq:obj} with time-sensitive parameter $\alpha_t$, $\beta_t$ and $\gamma_t$.
\end{itemize}
Furthermore, we also compared with the semi-supervised graph embedding baseline (i.e., GCN \cite{kipf2016semi}) to show the effectiveness of our proposed active learning query strategy on semi-supervised graph embedding baseline. For GCN, we follow the settings in their work by randomly sampling $(L+B)/C$ nodes for each class as the labelled training data.

As shown in Fig. \ref{fig:compareal}, by combining different node informativeness metrics (i.e, information entropy, density and graph centrality) and considering the time-sensitive parameters, our proposed AGE outperforms all the other baselines in terms of the node classification performance. Specifically, compared to the random baseline, AGE improves the node classification accuracy by 7.6\% (MacroF1) and 3.2\% (MicroF1) on Citeseer, 5.6\% (MacroF1) and 3.5\% (MicroF1) on Cora, 4.9\% (MacroF1) and 4.0\% (MicroF1) on Pubmed. Considering each AL criteria alone can only improve the performance to certain extend. Among the three AL criteria, information density is the most unstable one, which even brings a negative effect on Cora dataset. This explains why in the literature, representativeness based AL criterion (e.g., information density) is usually combined with the heterogeneity based criterion (e.g., information entropy). As illustrated in Fig. \ref{fig:compareal}, compared with the traditional ``Entropy$+$density based'' algorithm, involving graph centrality score improves the performance by  2\% in terms of MacroF1 and 0.9\% in terms of MicroF1 averagely. And as we analyzed in Section \ref{sec:ts}, by considering the time-sensitive parameters, the performance is further improved by 0.9\% (MacroF1) and 1.3\% (MicroF1) averagely. Pipeline approach does not provide satisfying performance. The reason may be that with the limited number of initial label nodes, GCN cannot embed the graph correctly. Then the nodes selected based on those node embedding may be not informative enough to provide sufficient information to train a good GCN. In our AGE framework, we select the nodes to label during the training of GCN. The two processes, active learning and graph embedding, reinforce each other during the training phase. Compared with the semi-supervised graph embedding baseline GCN, AGE achieves 0.3\% (MacroF1) and 2.2\% (MicroF1) improvements on Citeseer, 3.5\% (MacroF1) and 3.7\% (MicroF1) improvements on Cora, 3.4\% (MacroF1) and 3.7\% (MicroF1) improvements on Pubmed.

\section{Conclusions}
\label{sec:cl}
In this paper, we proposed a novel active learning framework for graph embedding named Active Graph Embedding (AGE). Unlike the traditional active learning algorithms, AGE processes the data with structural information and learnt representations (node embeddings), and it is carefully designed to address the challenges brought by these two characteristics. First, to exploit the graphical information, a graphical centrality based measurement is considered in addition to the popular information entropy based and information density based query criteria. Second, the active learning and graph embedding process are jointly run together by posing the label query at the end of every epoch of the graph embedding training process. Moreover, the time-sensitive weights are put on the three active learning query criteria which focus on the graphical centrality at the beginning and shift the focus to the other two embedding based criteria as the training process progresses (i.e., more accurate embeddings are learnt). We evaluate AGE on three public citation network datasets and verify the effectiveness of our framework design, including three query strategy criteria, time-sensitive parameters, by the node classification task. We further compare our proposed AGE with a pipeline baseline to show that active learning and graph embedding reinforce each other during the training process.

\bibliographystyle{ACM-Reference-Format}
\bibliography{sigproc} 

\end{document}